\pdfoutput=1

\documentclass[11pt]{article}

\usepackage[final]{emnlp2021}

\usepackage{times}
\usepackage{latexsym}

\usepackage[T1]{fontenc}

\usepackage[utf8]{inputenc}

\usepackage{microtype}

\usepackage{xspace}
\usepackage{hyperref}
\usepackage{graphicx}
\usepackage{multirow}
\usepackage{makecell}
\usepackage{booktabs}
\usepackage{algorithm}
\usepackage{algpseudocode}
\usepackage[caption=false]{subfig}
\algnewcommand\algorithmicforeach{\textbf{for each}}
\algdef{S}[FOR]{ForEach}[1]{\algorithmicforeach\ #1\ \algorithmicdo}

\usepackage{pifont}
\newcommand{\cmark}{\ding{51}}
\newcommand{\xmark}{\ding{55}}

\newcommand{\eg}{\textit{e.g., \xspace}}
\newcommand{\ie}{\textit{i.e., \xspace}}

\newcommand{\lib}{SummerTime\xspace}
\iffalse
    \newcommand{\an}[1]{\textcolor{cyan}{}}
\else
    \newcommand{\an}[1]{\textcolor{cyan}{\bf\small [AN: #1]}}
\fi

\title{\lib: Text Summarization Toolkit for Non-experts}

\author{
Ansong Ni$^\dagger$ 
\quad Zhangir Azerbayev$^\dagger$ 
\quad Mutethia Mutuma$^\dagger$ 
\quad Troy Feng$^\dagger$ \\
\bf \quad Yusen Zhang$^\clubsuit$ 
\bf \quad Tao Yu$^\dagger$ 
\bf \quad Ahmed Hassan Awadallah$^\diamondsuit$ 
\bf \quad Dragomir Radev$^\dagger$ \\ 
  $^\dagger$Yale University \quad
  $^\clubsuit$Penn State University \quad
  $^\diamondsuit$Microsoft Research \\
  \texttt{\{ansong.ni, tao.yu, dragomir.radev\}@yale.edu}}

\begin{document}
\maketitle
\begin{abstract}
Recent advances in summarization provide models that can generate summaries of higher quality. Such models now exist for a number of summarization tasks, including query-based summarization, dialogue summarization, and multi-document summarization. While such models and tasks are rapidly growing in the research field, it has also become challenging for non-experts to keep track of them. 
To make summarization methods more accessible to a wider audience, we develop \lib by rethinking the summarization task from the perspective of an NLP non-expert. \lib is a complete toolkit for text summarization, including various models, datasets and evaluation metrics, for a full spectrum of summarization-related tasks. 
\lib integrates with libraries designed for NLP researchers, and enables users with easy-to-use APIs.
With \lib, users can locate pipeline solutions and search for the best model with their own data, and visualize the differences, all with a few lines of code. We also provide explanations for models and evaluation metrics to help users understand the model behaviors and select models that best suit their needs. Our library, along with a notebook demo, is available at \url{https://github.com/Yale-LILY/SummerTime}.
\end{abstract}

\section{Introduction}
\begin{figure}
    \centering
    \includegraphics[width=\linewidth]{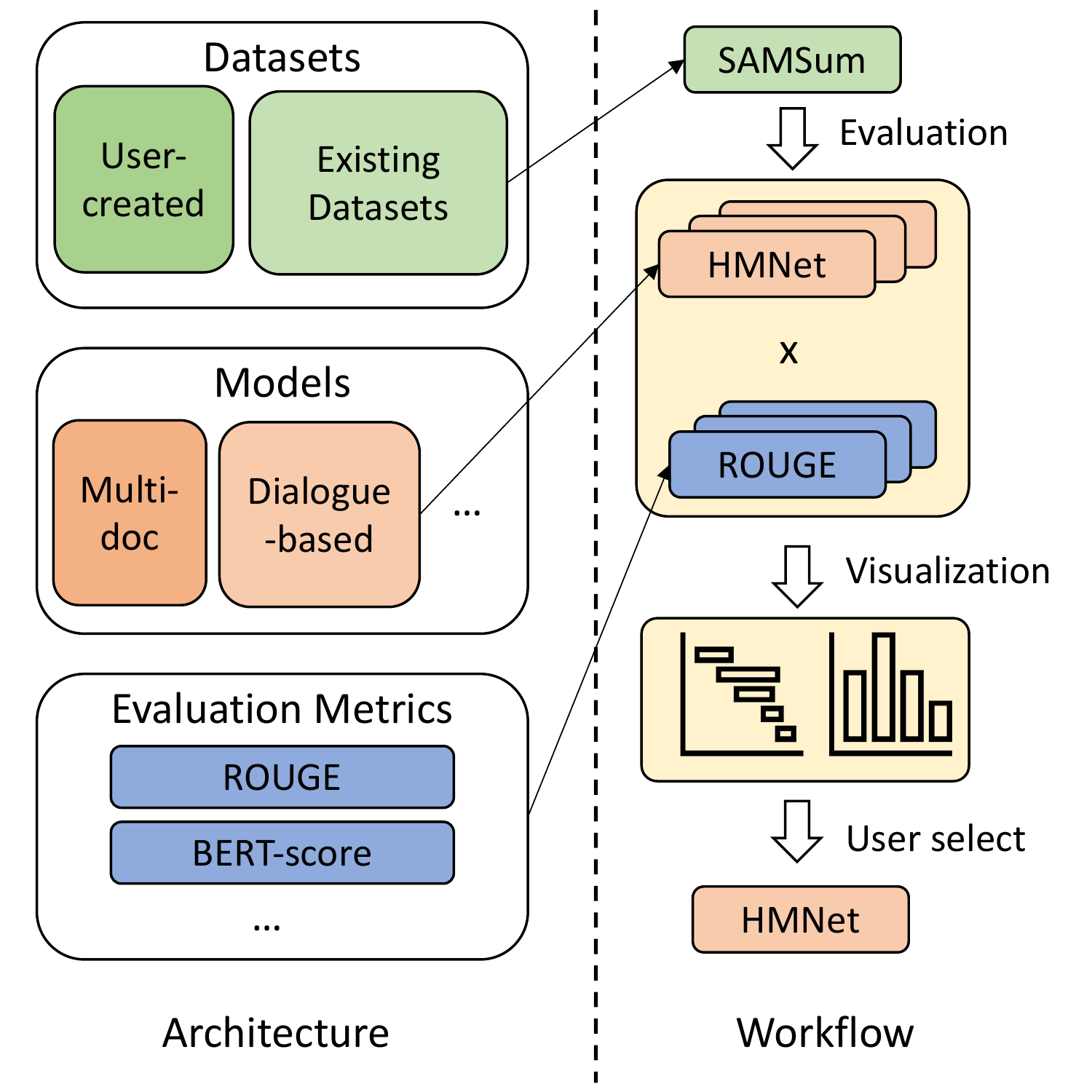}
    \caption{\lib is a toolkit for helping non-expert users to find the best summarization models for their own data and use cases.}
    \label{fig:arch}
\end{figure}
The goal of text summarization is to generate short and fluent summaries from longer textual sources, while preserving the most salient information in them. Benefiting from recent advances of deep neural networks, in particular sequence to sequence models, with or without attention ~\citep{sutskever2014sequence, bahdanau2014neural, vaswani2017attention}, current state-of-the-art summarization models  produce high quality summaries that can be useful in practice cases~\citep{zhang2020pegasus, lewis-etal-2020-bart}. 
Moreover, neural summarization has broadened its scope with the introduction of more summarization tasks, such as query-based summarization ~\citep{dang2005overview, zhong-etal-2021-qmsum}, long-document summarization~\citep{cohan-etal-2018-discourse}, multi-document summarization~\citep{ganesan2010opinosis, fabbri-etal-2019-multi}, dialogue summarization~\citep{gliwa-etal-2019-samsum, zhong-etal-2021-qmsum}. 
Such summarization tasks can also be from different domains \citep{hermann2015teaching, zhang2019optimizing, cohan-etal-2018-discourse}.

However, as the field rapidly grows, it is often hard for NLP non-experts to follow all relevant new models, datasets, and evaluation metrics. Moreover, those models and datasets are often from different sources, making it a non-trivial effort for the users to directly compare the performance of such models side-by-side. This makes it hard for them to decide which models to use. 
The development of libraries such as \textit{Transformers}~\citep{wolf-etal-2020-transformers} alleviate such problems to some extent, but they only cover a narrow range of summarization models and tasks and assume certain proficiency in NLP from the users, thus the target audience is still largely the research community.

To address those challenges for non-expert users and make state-of-the-art summarizers more accessible as a tool, we introduce \lib, a text summarization toolkit intended for users with no NLP background. We build this library from this perspective, and provide an integration of different summarization models, datasets and evaluation metrics, all in one place. We allow the users to view a side-by-side comparison of all classic and state-of-the-art summarization models we support, on their own data and combined into pipelines that fit their own task. SummerTime also provides the functionality for automatic model selection, by constructing pipelines for specific tasks first and iteratively evaluation to find the best working solutions.
Assuming no background in NLP, we list ``pros and cons'' for each model, and provide simple explanations for all the evaluation metrics we support. Moreover, we go beyond pure numbers and provide visualization of the performance and output of different models, to facilitate users in making decisions about which models or pipelines to finally adopt. 

The purpose of \lib is not to replace any previous work, on the contrary, we integrate  existing libraries and place them in the same framework. We provide wrappers around such libraries intended for expert users, maintaining the user-friendly and easy-to-use APIs.

\section{Related Work}
\subsection{Text Summarization}
Text summarization has been a long-standing task for natural language processing. Early systems for summarization had been focusing on extractive summarization \cite{mihalcea-tarau-2004-textrank, erkan2004lexrank}, by finding the most salient sentences from source documents. With the advancement of neural networks \citep{bahdanau2014neural, sutskever2014sequence}, the task of abstractive summarization has been receiving more attention \citep{rush2015neural, chopra2016abstractive, nallapati2016abstractive, celikyilmaz2018deep, chen2018fast, lebanoff2019scoring} while neural-based methods have also been developed for extractive summarization \citep{zhong2019closer, zhong2019searching, xu-durrett-2019-neural, cho2019improving, DBLP:conf/acl/ZhongLCWQH20, jia2020neural}. Moreover, the field of text summarization has also been broadening into several subcategories, such as multi-document summarization~\citep{mckeown1995generating, carbonell1998use, ganesan2010opinosis, fabbri-etal-2019-multi}, query-based summarization \citep{daume-iii-marcu-2006-bayesian, otterbacher2009biased, wang2016sentence, litvak-vanetik-2017-query, nema-etal-2017-diversity, baumel2018query, kulkarni2020aquamuse} and dialogue summarization \citep{zhong-etal-2021-qmsum, chen2021summscreen, chen2021dialsumm, gliwa-etal-2019-samsum, chen-yang-2020-multi, zhu-etal-2020-hierarchical}. The proposed tasks, along with the datasets can also be classified by domain, such as news \citep{hermann2015teaching, fabbri-etal-2019-multi, xsum-emnlp}, meetings \citep{zhong-etal-2021-qmsum, carletta2005ami, janin2003icsi}, scientifc literature \citep{cohan-etal-2018-discourse, yasunaga2019scisummnet}, and medical records \cite{deyoung2021ms2, zhang2019optimizing, portet2009automatic}.

\subsection{Existing Systems for Summarization}
\textit{Transformers} \citep{wolf-etal-2020-transformers} includes a large number of transformer-based models in its \textit{Modelhub}\footnote{https://huggingface.co/models}, including BART \citep{lewis-etal-2020-bart} and Pegasus \citep{zhang2020pegasus}, two strong neural summarizers we also use in \lib. It also hosts datasets for various NLP tasks in its \textit{Datasets}\footnote{https://huggingface.co/datasets} library \cite{lhoest2021datasets}. Despite the wide coverage in transformer-based models, \textit{Transformers} do not natively support models or pipelines that can handle aforementioned subcategories of summarization tasks. Moreover, it assumes certain NLP proficiency in its users, thus is harder for non-expert users to use. We integrate with \textit{Transformers} and \textit{Datasets} to import the state-of-the-art models, as well as summarization datasets into \lib, under the same easy-to-use framework. 

Another library that we integrate with is \textit{SummEval} \citep{fabbri2020summeval}, which is a collection of evaluation metrics for text summarization. \lib adopts a subset of such metrics in \textit{SummEval} that are more popular and easier to understand. \lib also works well with \textit{SummVis} \citep{vig2021summvis}, which provides an interactive way of analysing summarization results on the token-level. We also allow \lib to store output in a format that can be directly used by \textit{SummVis} and its UI. 

Other systems also exist for text summarization. \textit{MEAD}\footnote{http://www.summarization.com/mead/} is a platform for multi-lingual summarization. \textit{Sumy}\footnote{https://github.com/miso-belica/sumy} can produce extractive summaries from HTML pages or plain texts, using several traditional summarization methods including \citet{mihalcea-tarau-2004-textrank} and \citet{erkan2004lexrank}. \textit{OpenNMT}\footnote{https://github.com/OpenNMT/OpenNMT-py} is mostly for machine translation, but it also hosts several summarization models such as \citet{gehrmann2018bottom}.

\section{\lib}
The main purpose of \lib is to help non-expert users navigate through various summarization models, datasets and evaluation metrics, and provide simple yet comprehensive information for them to select the models that best suit their needs. \autoref{fig:arch} shows how \lib is split into different modules to help users achieve such goal. 

We will describe in detail each component of \lib in the following sections. With \autoref{sec:models}, we introduce the models we support in all subcategories of summarization; in \autoref{sec:datasets} we list all the existing datasets we support and how users can create their own evaluation set. Finally in \autoref{sec:evaluation}, we explain the evaluation metrics included with \lib and how they can help users find the most suitable model for their task.

\subsection{Summarization Models}
\label{sec:models}
Here we introduce the summarization tasks \lib covers and the models we include to support these tasks. We first introduce the single-document summarization models (\ie ``base models'') in \lib, and then we show how those models can be used in a pipeline with other methods to complete more complex tasks such as query-based summarization and multi-document summarization.

\subsubsection*{Single-document Summarization}
The following base summarization models are used in \lib. They all take a single document and generate a short summary. \\
\noindent\textbf{TextRank}~\citep{mihalcea-tarau-2004-textrank} is a graph-based ranking model that can be used to perform extractive summarization; \\
\noindent\textbf{LexRank}~\citep{erkan2004lexrank} is also a graph-based extractive summarization model, which is originally developed for multi-document summarization, but can also be applied to a single document. It uses centrality in a graph representation of sentences to measure their relative importance;\\
\noindent\textbf{BART}~\citep{lewis-etal-2020-bart} is an autoencoder model trained with denoising objectives during training. This seq2seq model is constructed with a bidirectional transformer encoder and a left-to-right transformer decoder, which can be fine-tuned to perform abstractive summarization; \\
\noindent\textbf{Pegasus}~\citep{zhang2020pegasus} proposes a new self-supervised pretraining objective for abstractive summarization, by reconstructing the target sentence with the remaining sentences in the document, it also shows strong results in low-resource settings; \\
\noindent\textbf{Longformer}~\citep{Beltagy2020Longformer} addresses the problem of memory need for self-attention models by using a combination of sliding window attention and global attention to approximate standard self-attention. It is able to support input length of 16K tokens, a large improvement over previous transformer-based models.

\subsubsection*{Multi-document Summarization}
For multi-document summarization, we adopt two popular single-document summarizers to complete the task, as this is shown to be effective in previous work \citep{fabbri-etal-2019-multi}. \\
\noindent\textbf{Combine-then-summarize} is a pipeline method to handle multiple source documents, where the documents are concatenated and then a single document summarizer is used to produce the summary. Note that the length of the combined documents may exceed the input length limit for typical transformer-based models; \\
\noindent\textbf{Summarize-then-combine} first summarizes each source document independently, then merges the resulting summaries. Compared to the combine-then-summarize method, it is not affected by overlong inputs. However, since each document is summarized separately, the final summary may contain redundant information \citep{carbonell1998use}.

\subsubsection*{Query-based Summarization}
For summarization tasks based on queries, we adopt a pipeline method and first use retrieval methods to identify salient sentences or utterances in the original document or dialogue, then generate summaries with a single-document summarization model. \\
\noindent\textbf{TF-IDF retrieval} is used in a pipeline to first retrieve the sentences that are most similar to the query based on the TF-IDF metric; \\
\noindent\textbf{BM25 retrieval} is used in the same pipeline, but BM25 is used as the similarity metric for retrieving the top-$k$ relevant sentences.

\subsubsection*{Dialogue Summarization}
Dialogue summarization is used to extract salient information from a dialogue. \lib includes two methods for dialogue summarization. \\
\noindent\textbf{Flatten-then-summarize} first flattens the dialogue data while preserving the speaker information, then a summarizer is used to generate the summary. \citet{zhong-etal-2021-qmsum} found that this presents a strong baseline for dialogue summarization. \\
\noindent\textbf{HMNet}~\cite{zhu-etal-2020-hierarchical} explores the semantic structure of dialogues and develops a hierarchical architecture to first encode each utterance then aggregate with another encoder in modeling the long dialogue script. It also exploits role vectors to perform better speaker modeling. 

\begin{figure}
    \centering
    \includegraphics[width=\linewidth]{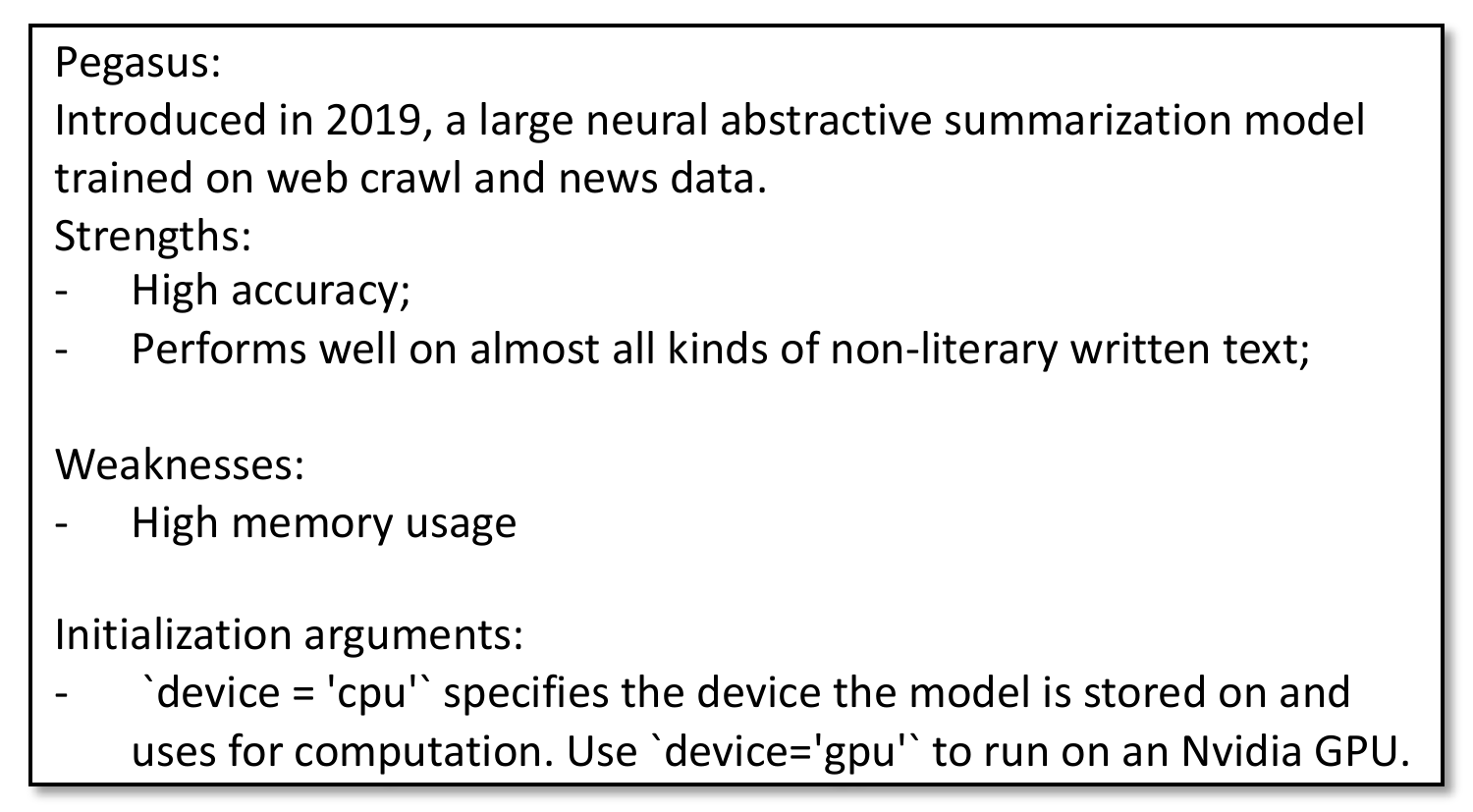}
    \caption{A short description of the Pegasus model, \lib includes such short descriptions for each supported models to help user making choices.}
    \label{fig:model-description}
\end{figure}

Since we assume no NLP background of our target users, we provide a short description for every model to illustrate the strengths and weaknesses for each model. Such manually written descriptions are displayed when calling a static \texttt{get\_description()} method on the model class. A sample description is shown in \autoref{fig:model-description}.

\subsection{Datasets}
\label{sec:datasets}
\begin{table*}[t]
    \centering
    \small
    \begin{tabular}{l|ccccccccc}
    \toprule
        Dataset & Domain & Size & Src. length & Tgt. length & Query & Multi-doc & Dial. & Lang.\\\midrule
        CNN/DM(3.0.0) & News & 300k & 781 & 56 &  \xmark & \xmark & \xmark & En \\
        Multi-News & News & 56k & 2.1k & 263.8 & \xmark & \cmark & \xmark & En \\
        SAMSum & Open-domain & 16k & 94 & 20 & \xmark & \xmark & \cmark & En \\
        XSum & News & 226k & 431 & 23.3 & \xmark & \xmark & \xmark & En \\
        ScisummNet & Scientific articles & 1k & 4.7k &  150 & \xmark & \xmark & \xmark & En \\
        QMSum & Meetings & 1k & 9.0k & 69.6 & \cmark & \xmark & \cmark & En \\
        ArXiv & Scientific papers & 215k & 4.9k & 220 & \xmark & \xmark & \xmark & En \\
        PubMedQA & Biomedial & 273.5k & 239 & 43 & \cmark & \xmark & \xmark & En \\
        SummScreen & TV shows & 26.9k & 6.6k & 337.4 & \xmark & \xmark & \cmark & En \\
        MLSum & News & 1.5M & 635 & 31.8 & \xmark & \xmark & \xmark & \makecell{Fr, De, Es, \\ Ru, Tr} \\
    \bottomrule
    \end{tabular}
    \caption{The summarization datasets included in \lib. ``Dial.'' is short for ``Dialogue'' while ``Lang.'' denotes the languages of each of the datasets.}
    \label{tab:datasets}
\end{table*}

With \lib, users can easily create or convert their own summarization datasets and evaluate all the supporting models within the framework. However, in the case that no such datasets are available, \lib also provides access to a list of existing summarization datasets. This way, users can select models that perform the best on one or more datasets that are similar to their task.

\noindent\textbf{CNN/DM}~\citep{hermann2015teaching} contains news articles from CNN and Daily Mail. Version 1.0.0 of it was originally developed for reading comprehension and abstractive question answering, then the extractive and abstractive summarization annotations were added in version 2.0.0 and 3.0.0, respectively; \\
\noindent\textbf{Multi-News}~\citep{fabbri-etal-2019-multi} is a large-scale multi-document summarization dataset which contains news articles from the site newser.com with corresponding human-written summaries. Over 1,500 sites, i.e. news sources, appear as source documents, which is higher than the other common news datasets. \\
\noindent\textbf{SAMSum}~\citep{gliwa-etal-2019-samsum} is a dataset with chat dialogues corpus, and human-annotated abstractive  summarizations. In the SAMSum corpus, each dialogue is written by one person. After collecting all the dialogues, experts write a single summary for each dialogue. \\
\noindent\textbf{XSum}~\citep{xsum-emnlp} is a news summarization dataset for generating a one-sentence summary aiming to answer the question “What is the article about?”. It consists of real-world articles and corresponding one-sentence summarization from British Broadcasting Corporation (BBC). \\
\noindent\textbf{ScisummNet}~\citep{yasunaga2019scisummnet} is a human-annotated dataset made for citation-aware scientific paper summarization (Scisumm). It contains over 1,000 papers in the ACL anthology network as well as their citation networks and their manually labeled summaries. \\
\noindent\textbf{QMSum}~\citep{zhong-etal-2021-qmsum} is designed for query-based multi-domain meeting summarization. It collects the meetings from AMI and ICSI dataset, as well as the committee meetings of the Welsh Parliament and Parliament of Canada. Experts manually wrote summaries for each meeting. \\
\noindent\textbf{ArXiv}~\citep{cohan-etal-2018-discourse} is a dataset extracted from research papers for abstractive summarization of single, longer-form documents. For each research paper from arxiv.org, its abstract is used as ground-truth summaries. \\
\noindent\textbf{PubMedQA}~\citep{jin-etal-2019-pubmedqa} is a question answering dataset on the biomedical domain. Every QA instance contains a short answer and a long answer, latter of which can also be used for query-based summarization. \\
\noindent\textbf{SummScreen}~\citep{chen2021summscreen} consists of community contributed transcripts of television show episodes from The TVMegaSite, Inc. (TMS) and ForeverDream (FD). The summary of each transcript is the recap from TMS, or a recap of the FD shows from Wikipedia and TVMaze. \\
\noindent\textbf{MLSum}~\citep{scialom-etal-2020-mlsum} is a large-scale multilingual summarization dataset. It contains over 1.5M news articles in five languages, namely French, German, Spanish, Russian, and Turkish.

A summary of all datasets included in \lib is shown as \autoref{tab:datasets}, it is worth noticing that the fields in this table (\ie domain, query-based, multi-doc, etc) are also incorporated in each of the dataset classes (\eg \texttt{SAMSumDataset}) as class variables, so that such labels can later be used to identify applicable models. Similar with the models classes, we include a short description for each of the datasets.  Note that the datasets, either existing ones or user created are mainly for evaluation purposes. We leave the important task of fine-tuning the models on these datasets for future work, for which we describe in more detail in \autoref{sec:future-work}.

\subsection{Evaluation Metrics}
\label{sec:evaluation}
To evaluate the performance of each supported model on certain dataset, \lib integrates with \textit{SummEval}~\citep{fabbri2020summeval} and provides the following evaluation metrics for the users to understand model performance: \\
\noindent\textbf{ROUGE}~\citep{lin2004rouge} is a recall-oriented method based on overlapping n-grams, word sequences, and word pairs between the generated output and the gold summary; \\
\noindent\textbf{BLEU}~\citep{papineni2002bleu} measures n-gram precision and employs a penalty for brevity, BLEU is often used as an evaluation metric for machine translation; \\
\noindent\textbf{ROUGE-WE}~\citep{ng-abrecht-2015-better} aims to go beyond surface lexical similarity and uses pretrained word embeddings to measure the similarity between different words and presents a better correlation with human judgements; \\
\noindent\textbf{METEOR}~\citep{lavie2007meteor} is based on word-to-word matches between generated and reference summaries, it consider two words as ``aligned'' based on a Porter stemmer~\citep{porter2001snowball} or synonyms in WordNet~\citep{miller1995wordnet}; \\
\noindent\textbf{BERTScore}~\citep{bert-score} computes token-level similarity between sentences with the contextualized embeddings of each tokens.

Since we assume no NLP background from our target users, we make sure that \lib provides a short explanation for each evaluation metric as well as a clarification whether high or low scores are better for a given evaluation metric, to help the non-expert users understand the meaning of the metrics and use them to make decisions. 

\begin{figure}
    \centering
    \includegraphics[width=\linewidth]{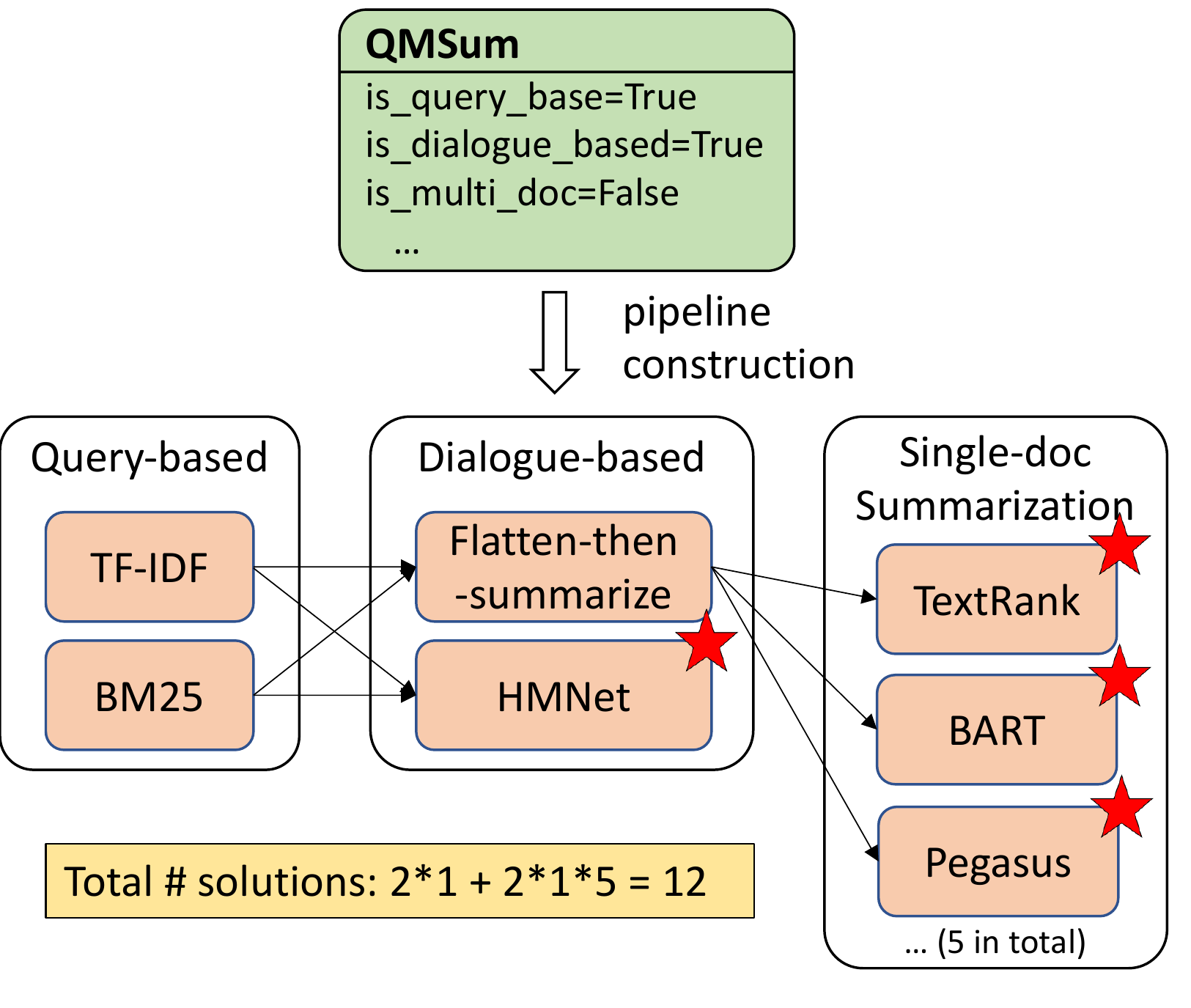}
    \caption{An illustration of how \lib finds solutions to a specific tasks defined by a dataset. The red star denotes that an ending point is reached.}
    \label{fig:pip-const}
\end{figure}
\begin{algorithm}[!t] 
\small
\caption{\textsc{Select}($\mathcal M, \mathcal D, \mathcal E$)}
\label{alg:model-selection}
\begin{algorithmic}[1]
\Require $\mathcal M$: a pool of models to choose from, 
$\mathcal D$: a set of examples from a dataset, 
$\mathcal T$: a set of evaluation metrics, $d$: initial resource number, $k$: increase resource factor
\Ensure $M \subseteq \mathcal{M}$: a subset of models; \\
Initialize $M=\mathcal{M}, M'=\emptyset$
\While {$M' \neq M$}
\State $D = sample(\mathcal{D}, d)$
\ForEach {$m \in M, e \in \mathcal E $}
\State $r_m^e = eval(m, D, e)$
\EndFor
\State $M'=M$
\ForEach{$m \in M$}
\If {$\exists m'$ s.t. $r^e_{m'}>r^e_m, \forall e \in \mathcal{E}$}
\State $M = M\backslash m$
\EndIf
\EndFor
\State $d=d*k$
\EndWhile
\end{algorithmic}
\end{algorithm} 

\section{Model Selection}
\begin{figure}[ht]
    \centering
    \includegraphics[width=\linewidth]{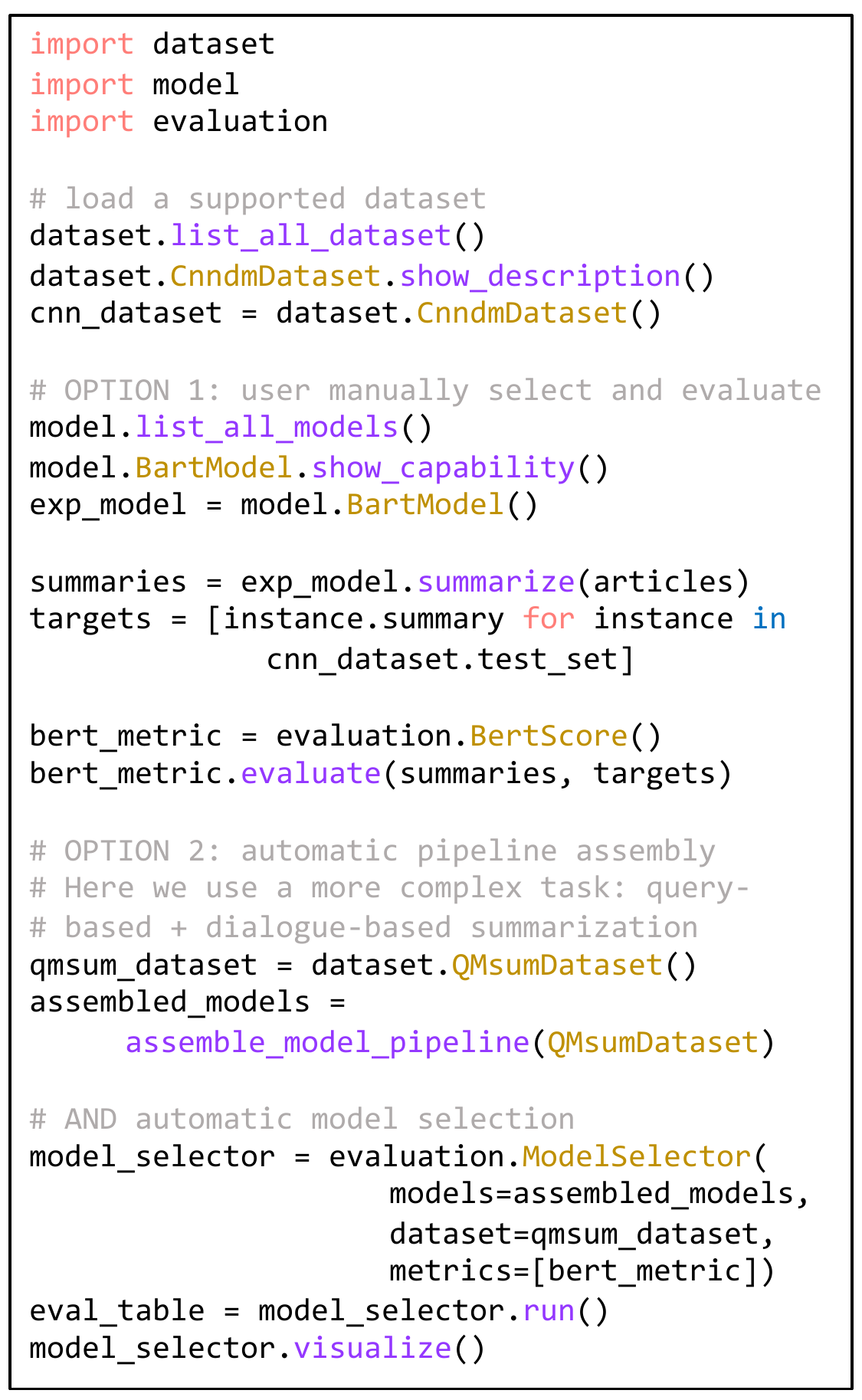}
    \caption{Example code for using \lib. Additionally, we show two ways for performing model selection and evaluation.}
    \label{fig:code-example}
\end{figure}

In this section, we describe in detail about the workflow of \lib and how it can help our non-expert users find the best models for their use cases, which is one of the main functionalities that makes \lib stands out from similar libraries. A concrete code example of this is shown in \autoref{fig:code-example}.

\paragraph{Create/select datasets} The user would first load a dataset with the APIs we provide. During the process, the users also need to specify some Boolean attributes (\eg \texttt{is\_query\_based}, \texttt{is\_dialogue\_based}) to facilitate next steps. Alternatively, the user can also choose to use one of the datasets that are included in \lib, where such attributes are already specified in \autoref{tab:datasets}. 
\paragraph{Construct pipelines} After identifying the potential pipeline modules (\eg query-based module, dialogue-based module) that are applicable to the task, \lib automatically constructs solutions to a specific dataset by combining the pipelines and summarization models specified in \autoref{sec:models}. It further places all such constructed solutions in a pool for further evaluation and selection purposes. 
An example of this process in shown in \autoref{fig:pip-const}.
\paragraph{Search for the best models} 
As shown in \autoref{fig:pip-const}, there can be a large pool of solutions to be evaluated. To save time and resources in searching for best models, \lib adopts the idea of successive halving~\citep{li2017hyperband, jamieson2016non}. More specifically, \lib first uses a small number of examples from the dataset to evaluate all the candidates and eliminate models that are surpassed by at least one other model on every evaluation metric, then it does so iteratively and gradually increases the evaluation set size to reduce the variance. As shown in \autoref{alg:model-selection}, the final output is a set of competing models $M$ that are better\footnote{Note that in line 9 of the algorithm, the symbol ``$>$'' is conceptual and should be interpreted as ``better than''} than one another on at least one metric.

\paragraph{Visualization}
\begin{figure}[ht]
\centering
\subfloat[Visualize the performance distribution of the models over the examples.]{%
  \includegraphics[clip,width=0.9\linewidth]{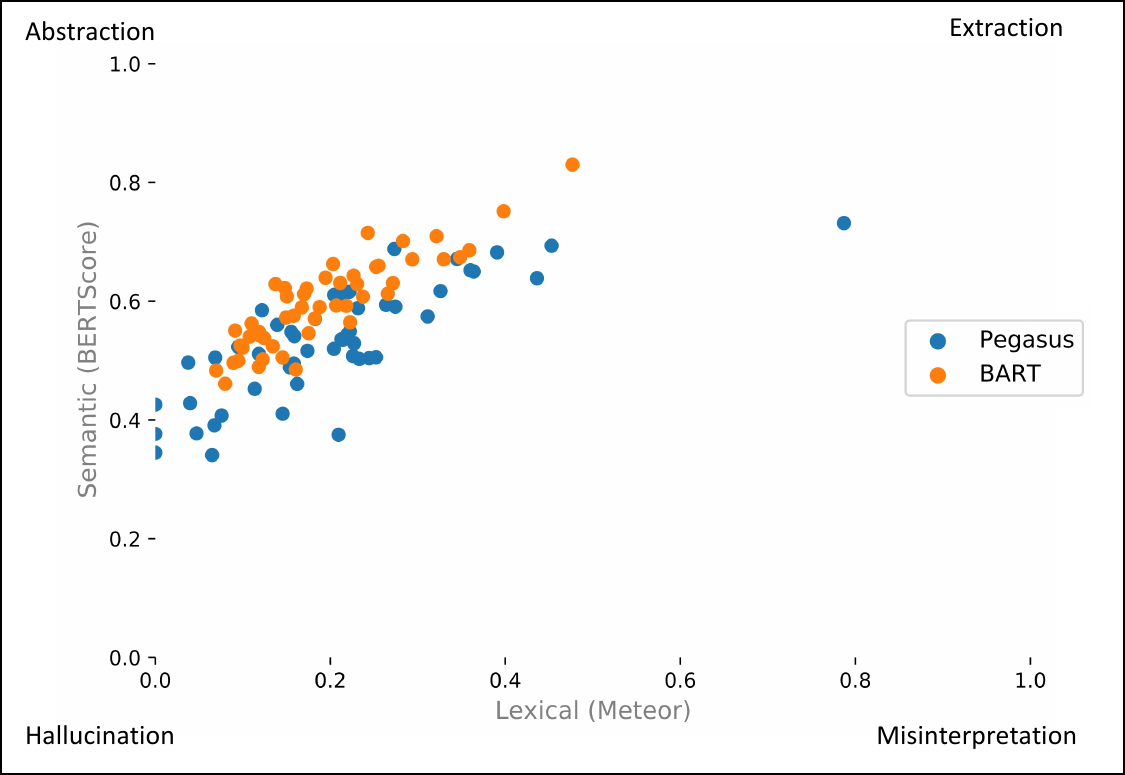}%
}

\subfloat[Visualize the performance of models over different evaluation metrics.]{%
  \includegraphics[clip,width=0.8\linewidth]{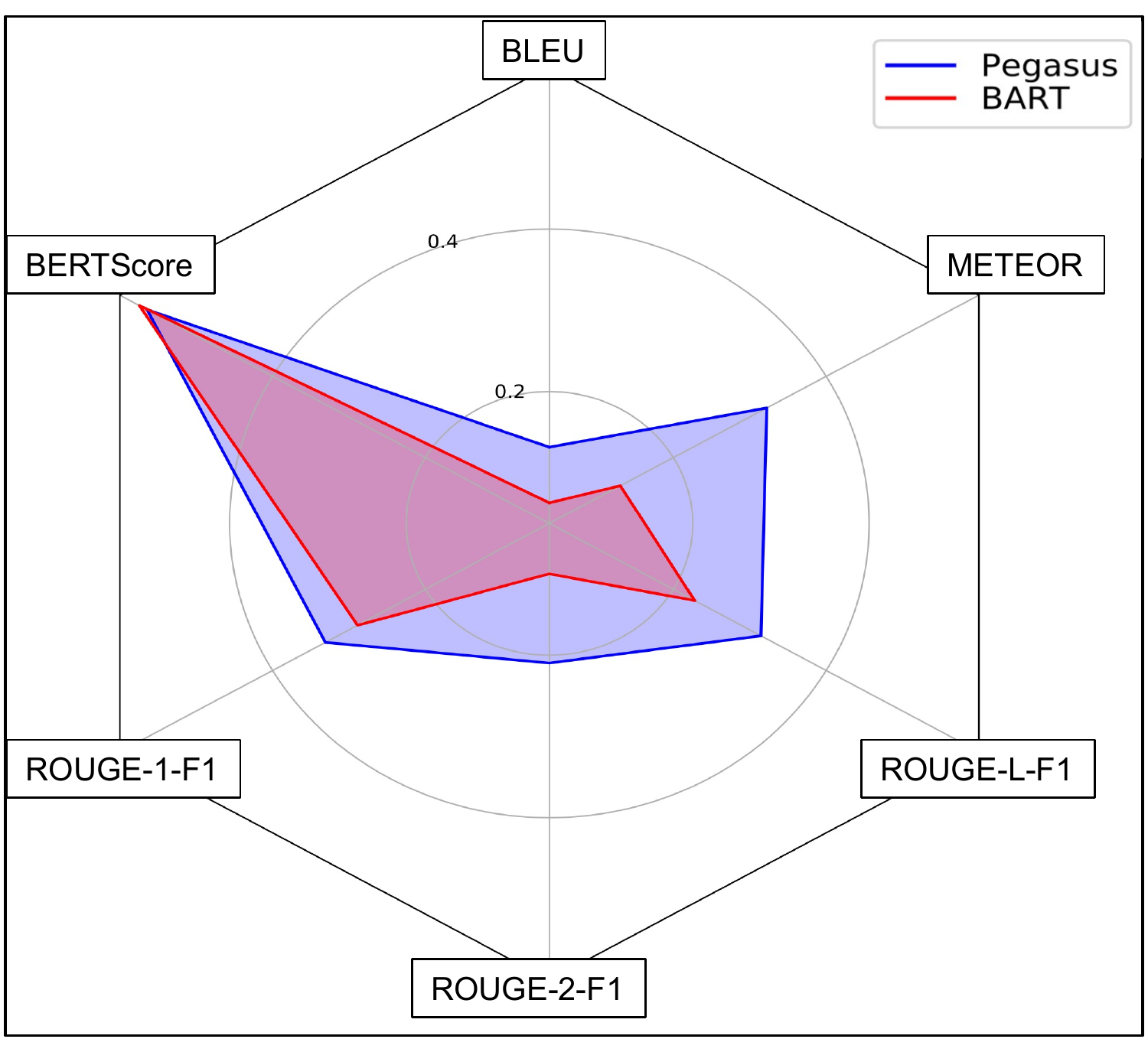}%
}
\caption{Examples of the visualization \lib provides for the users to better compare the performance between different models.}
\label{fig:plots}
\end{figure}
In addition to showing the numerical results as tables, \lib also allows the users to visualize the differences between different models with different charts and \textit{SummVis} \citep{vig2021summvis}.
\autoref{fig:plots} shows some examples of such visualization methods \lib provides. A scatter plot can help the users understand the distribution of the model's performance over each example, while the radar chart is an intuitive way of comparing different models over various metrics.
\lib can also output the generated summaries to file formats that are directly compatible with \textit{SummVis} \cite{vig2021summvis}, so that the users can easily use it to visualize the per-instance output differences on the token level.

\section{Future Work}
\label{sec:future-work}
An important piece of future work for \lib is to include more summarization models (\eg multilingual, query-aware, etc) to enlarge the number of choices for the users, and more datasets to increase the chance of users finding similar tasks or domain for evaluation when they do not have a dataset of their own. 
We also plan to add more visualization methods for the users to better understand the differences between the outputs of various models and the behavior of each individual model itself. 
Moreover, we would like to enable fine-tuning for a subset of smaller models we support, to enable better performance on some domains or tasks for which no pretrained models are available. With all such potential improvements in the near future, we plan to supply SummerTime not only as a way for non-expert users to access state-of-art summarization models, but also as a go-to choice to quickly establishing baseline results for researchers as well. 

\section{Conclusion}
We introduce \lib, a text summarization toolkit designed for non-expert users. \lib includes various summarization datasets, models and evaluation metrics and covers a wide range of summarization tasks. It can also automatically identify the best models or pipelines for a specific dataset and task, and visualize the differences between the model outputs and performances. \lib is open source under the Apache-2.0 license and is available online. 

\section*{Acknowledgements}
The authors would like to thank Rui Zhang, Alexander Fabbri, Chenguang Zhu, Budhaditya Deb, Asli Celikyilmaz and Rahul Jha for their advice for this work. This work is supported in part by a grant from Microsoft Research.

\bibliography{anthology,custom}
\bibliographystyle{acl_natbib}




\end{document}